\documentclass[letterpaper, 10 pt, conference]{ieeeconf}  % Comment this line out if you need a4paper

\IEEEoverridecommandlockouts                              % This command is only needed if 
\usepackage{graphics} % for pdf, bitmapped graphics files
\usepackage{epsfig} % for postscript graphics files

\usepackage{amsmath} % assumes amsmath package installed
\usepackage{amssymb}  % assumes amsmath package installed
\usepackage{caption}
\usepackage{subfig}
\usepackage{multirow}
\usepackage{color}

\usepackage{amssymb, eucal, xfrac}
\usepackage[margin=6pt,font=small,labelfont=bf,labelsep=endash,tableposition=top]{caption}

\def\spadesuit{\dag}
\def\spadesuit{\dag}
\def\clubsuit{\ddag}

\usepackage{xcolor}		
\usepackage[colorlinks = true,
            linkcolor = purple,
            urlcolor  = blue,
            citecolor = cyan,
            anchorcolor = black]{hyperref}

\title{\LARGE \bf FastFlowNet: A Lightweight Network for Fast Optical Flow Estimation
}

\author{Lingtong Kong$^{\spadesuit}$, 
~ ~ ~  Chunhua Shen$^{\clubsuit}$, 
 ~ ~ ~ Jie Yang$^{\spadesuit,*}$% <-this % stops a space
\thanks{$^\spadesuit$ Institute of Image Processing and Pattern Recognition, Shanghai Jiao Tong University, China. $^\clubsuit$ The University of Adelaide, Australia.}
\thanks{$^*$ Corresponding author.
}
}

\begin{document}

\maketitle
\thispagestyle{empty}
\pagestyle{empty}

%%%%%%%%%%%%%%%%%%%%%%%%%%%%%%%%%%%%%%%%%%%%%%%%%%%%%%%%%%%%%%%%%%%%%%%%%%%%%%%%
\begin{abstract}

    Dense optical flow estimation plays a key role in many robotic vision tasks. 
    % It has been predicted with satisfying accuracy than traditional methods with advent of deep learning.
    In the past few years, 
    with the advent of deep learning,  we have witnessed great progress in optical flow estimation. 
    However, current networks often %occupy
    consist of 
    a large number of parameters and require heavy computation costs, 
    %These drawbacks have hindered applications on power- or memory-constrained mobile devices. 
    largely hindering its application on low power-consumption devices such as mobile phones.
    % To deal with these challenges, in this paper,
    In this paper, we tackle this challenge and design a lightweight model
    %we
    %by diving into designing efficient structure 
    for fast and accurate optical flow prediction.
    Our proposed FastFlowNet %works in the well-known 
    follows the widely-used
    \textit{coarse-to-fine} %manner
    paradigm
    with following innovations. First, a new head enhanced pooling pyramid (HEPP) feature extractor is employed to intensify high-resolution pyramid features while reducing parameters. Second, we introduce a new center dense dilated correlation (CDDC) layer for constructing compact cost volume that can keep large search radius with reduced computation burden. Third, an efficient shuffle block decoder (SBD) is implanted into each pyramid level to accelerate flow estimation with marginal drops in accuracy. Experiments on both synthetic Sintel data and real-world KITTI datasets demonstrate the effectiveness of the proposed approach, which %consumes
    needs 
    only $ \sfrac{1}{10}$  computation of comparable networks to %get
    achieve 
    % $90\%$  of their
    % performance. 
    \textit{on par} accuracy.  
    In particular, FastFlowNet only contains 1.37M parameters; and can
    %runs
    execute 
    at 90 FPS (with a single GTX 1080Ti) or 5.7 FPS (embedded Jetson TX2 GPU)
    %with  or 
    on a pair of Sintel images of resolution  $1024 \times  436$.

Code %and models are
is 
available at: 
\def\UrlFont{\tt \color{blue}}
\url{https://git.io/fastflow}

% https://github.com/ltkong218/FastFlowNet

\end{abstract}

%%%%%%%%%%%%%%%%%%%%%%%%%%%%%%%%%%%%%%%%%%%%%%%%%%%%%%%%%%%%%%%%%%%%%%%%%%%%%%%%
\section{Introduction}

Optical flow estimation is a fundamental task in computer vision and 
%in
benefits 
a variety of downstream vision tasks, including target tracking~\cite{9093517}, autonomous navigation~\cite{8460569}, obstacle avoidance \cite{Kahlouche_2007} and action-based human-robot interaction~\cite{8793825,Yang_2021}. Given two time-adjacent image frames,  optical flow
offers rich information for robotics to interact in complex dynamic environments,
by estimating the projected 2D velocity field on the image plane, which is caused by relative 3D motion between an observer and a scene.  
%With the popularization of smart mobile devices, image sensors and embedded processors have achieved much better performance with a more afforadable price, making vision based motion estimation cost effective in robotics. 
However, extracting  % meaningful 
accurate 
motion information from %raw camera data
RGB images 
is a complicated and computationally intensive task.
%where 
Decades of research efforts have been %made.
spent on optical flow estimation. 
While traditional methods~\cite{Horn_1981,Brox_2011,Zach_2007,Kroeger_2016} of optimizing an energy function with brightness constancy and spatial smoothness usually fail in large movement and illumination change cases, recent deep learning approaches significantly surpass them in both accuracy and %running 
speed thanks to large synthetic datasets and powerful GPUs. Existing Convolutional Neural Networks (CNNs) based architectures can be categorized into two classes:
%one is 
the 
encoder-decoder structure and the %other is
coarse-to-fine %residual
structure.

\begin{figure}[t]
	\centering
	\vspace{0.8 em}
	\includegraphics[width=0.45\textwidth]{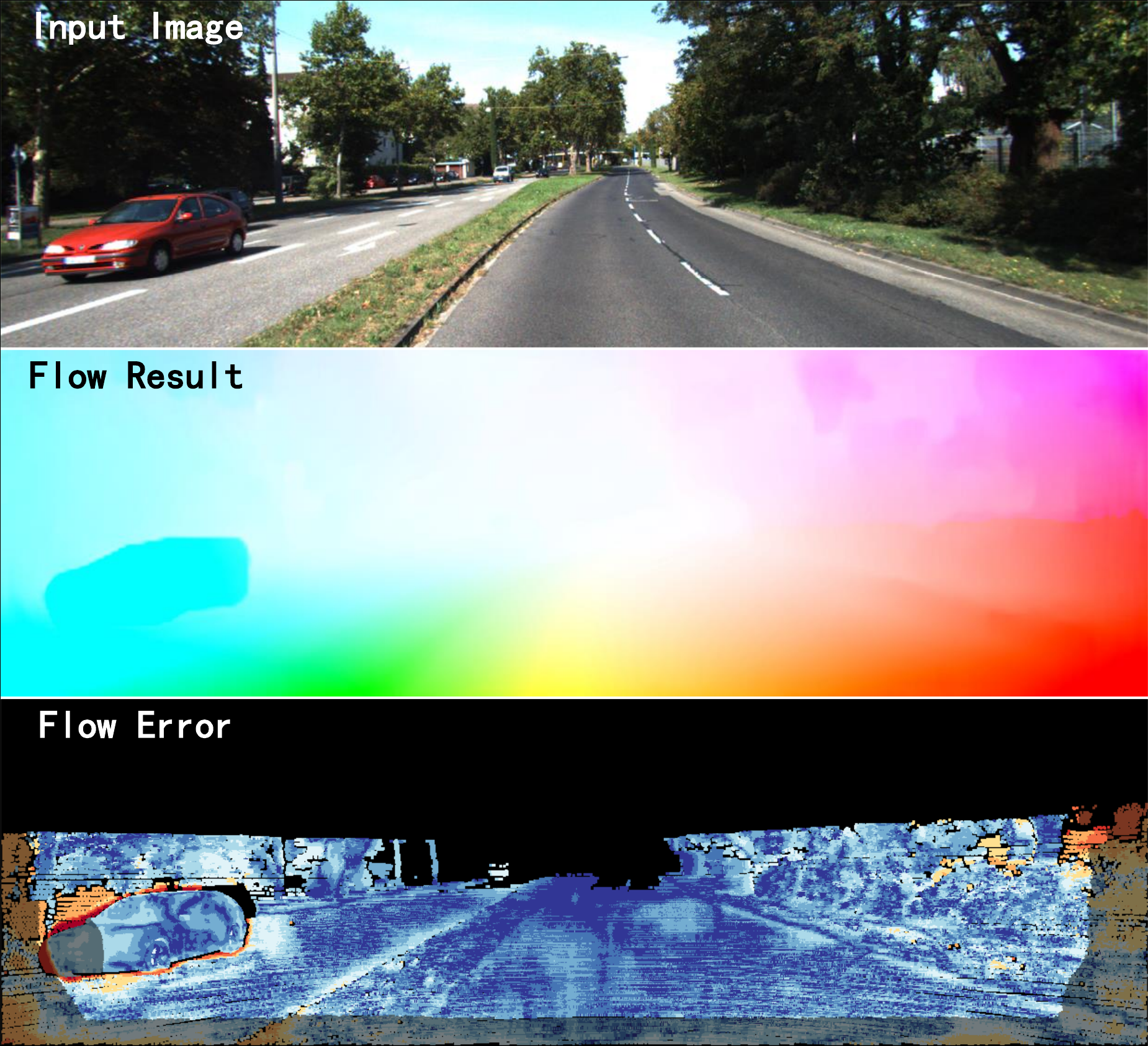}
	\caption{FastFlowNet can estimate accurate optical flow in real-time on embedded devices, which consumes only $ \sfrac{1}{10} $ computation of comparable networks to achieve \textit{on par} performance.}
	\label{fig1}
\end{figure}

%
% XXXXXX
%
Representative works of the first %class
category 
% are
include  
FlowNet~\cite{Fischer_2015} and FlowNet2 \cite{Ilg_2017}. The pioneering FlowNet~\cite{Fischer_2015} %puts up
designs  
two versions of models,  named FlowNetS and FlowNetC. They both adopt the U-shape network. Specifically, FlowNetS only concatenates two input images while FlowNetC also contains a correlation layer. The successor FlowNet2~\cite{Ilg_2017} further cascades multiple FlowNet models and 
% provides
employs 
carefully designed learning schedules on multiple datasets. % They have shown that
FlowNet2 shows that 
CNNs can learn to estimate accurate optical flow with orders of magnitude faster than traditional competitors. However, these networks %have to keep
typically use a 
large number of channels in low resolutions for feature encoding, resulting in %big
large
model sizes, \textit{e.g.}, FlowNet2 containing more than 160M parameters. Moreover, the correlation layer in FlowNetC locates at a relative high resolution with large search radius, which leads to heavy computation burden. Both %defects
shortcomings  
have hindered them for lightweight mobile applications.

On the other hand, some approaches %try
attempt 
to estimate residual flow among decomposed spatial levels where SPyNet~\cite{Ranjan_2017} 
may be the first one  % beginning.
in this category.
SPyNet  first constructs image pyramids, then concatenates the upsampled prior flow, first image and warped second image to estimate residual flow fields in a coarse-to-fine manner. Since each sub-module is only responsible for the small magnitude residual flow, SPyNet~\cite{Ranjan_2017} can % approach
achieve the 
accuracy of FlowNet~\cite{Fischer_2015} with only 1.2M parameters. 
% CS: THIS IS NOT VERY ACCURATE
% The main shortcomings are obvious performance degradation and large computation cost at high resolutions. 
%
Almost at the same time, PWC-Net~\cite{Sun_2018_CVPR} and LiteFlowNet~\cite{Hui_2018_CVPR} replace image pyramids  with better feature pyramids  and introduce the correlation layer into each spatial level for better correspondence representation. Highly ranked results confirm the effectiveness of coarse-to-fine based approaches.
% where great attention has been paid in following
As a result, following works appeared along this line of research \cite{Hur_2019, Yin_2019, zhao2020maskflownet, Kong_2020, hofinger2019improving}.
%\cite{Kong_2021}.
% Nevertheless,

Most recent works on optical flow \cite{zhao2020maskflownet, hofinger2019improving, teed2020raft} have focused almost exclusively on pursuing accuracy, the resulting methods with intensive computation  %algorithms
are hardly deployed %on
to 
energy-constrained embedded devices, such as unmanned aerial vehicles (UAV)
and mobile phones.
%and 
%micro intelligent robots.
% 
% CS: THIS IS NOT A VALID ARG
% Also, their large number of parameters will occupy lots of storage space and increase the risk of overfitting.
%
% 
For example, as we 
show %that
later, 
the efficient LiteFlowNet~\cite{Hui_2018_CVPR} barely breaks the 1 FPS barrier when inferring Sintel resolution images ($1024 \times 436$) on Jetson TX2, which is still far from practical deployment.

To speed up accurate optical flow estimation and facilitate practical applications, we propose a lightweight and fast network,
%dubbed
termed 
\textbf{FastFlowNet} in this paper. Our model is based on the widely used coarse-to-fine residual structure and we improve it on three aspects: %that is
1) 
\textit{pyramid feature extraction}, 2) \textit{cost volume construction} and 3) \textit{optical flow decoding}, covering all components of flow estimation pipeline. 

\begin{figure*}[t]
	\centering
	\includegraphics[width=.85\textwidth]{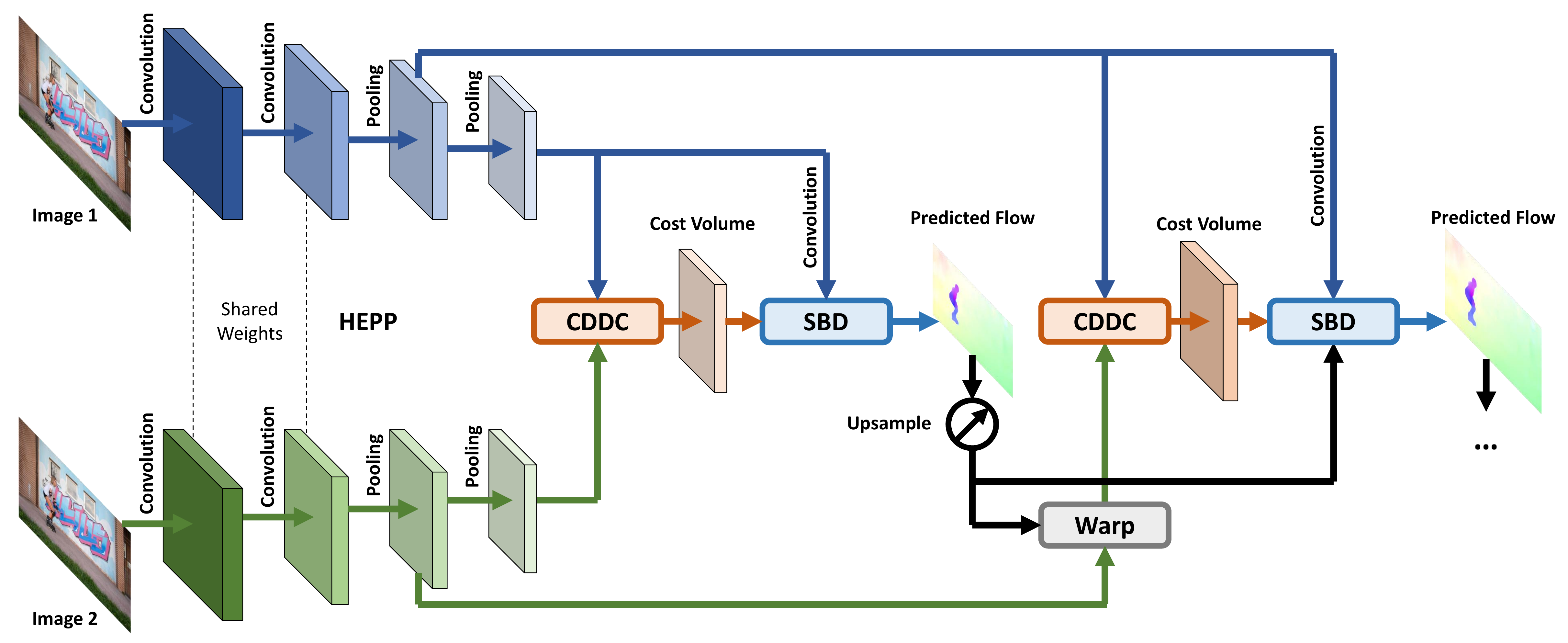}
	\caption{Architecture overview of the proposed \textbf{FastFlowNet}. HEPP, CDDC and SBD are efficient modules for extracting pyramid feature, constructing cost volume and regressing optical flow respectively. Only two levels are shown here for clarity.}
	\label{fig2}
\end{figure*}

First, we present a head enhanced pooling pyramid (HEPP) feature extractor, which uses convolution layers %in
at 
higher levels while adopting parameter-free pooling %in
at 
other lower levels. This efficient module can be %regarded
viewed 
as a combination of feature pyramid in PWC-Net~\cite{Sun_2018_CVPR} and pooling pyramid of SPyNet~\cite{Ranjan_2017}, %that
which 
aims to extract %relative
good matching features  with  % obviously
considerably 
reduced parameters and computation. Second, recent studies \cite{Sun_2018_CVPR, hofinger2019improving} have shown that increasing the  search radius of the correlation layer can improve flow accuracy. However, feature channels of the cost volume are squared in terms of the search radius and computation complexity of following decoder network %will be fourth power
is  4th power 
of the search radius. In FastFlowNet, we keep the search radius to be 4 as in  PWC-Net~\cite{Sun_2018_CVPR} for perceiving large movement. Differently, to reduce computation burden, we down sample feature channels in large residual regions and propose a %novel
new 
center dense dilated correlation (CDDC) layer for constructing compact cost volume. Our motivation is that residual flow distributions are more focused on small motions, and experiments show that CDDC behaves better than other compression methods. Third, we observe that flow decoders %in
at 
each pyramid level occupy a relatively large proportion of parameters and computation as for the whole network. To further reduce computation and meanwhile %hold
retain 
superior performance, we build a new shuffle block decoder (SBD) module referring to the lightweight ShuffleNet~\cite{8578814}, for its low computation budget and high classification precision. Different from ShuffleNet~\cite{8578814} as a backbone network, our SBD module is employed for regressing optical flow, %that
which 
is only located in the middle part of decoder network. The overall architecture of FastFlowNet is % depicted
illustrated 
in Fig.~\ref{fig2}.

Based on the above improvements, our proposed network can achieve % advanced
impressive 
performance on the Sintel~\cite{Butler_2012} and KITTI~\cite{7298925} benchmarks with significantly reduced computation %budget.
cost. 
In particular, FastFlowNet runs $3\times$ faster than PWC-Net~\cite{Sun_2018_CVPR} with $\sfrac{1}{7}$ computation; and $5\times$ faster than LiteFlowNet~\cite{Hui_2018_CVPR} with $\sfrac{1}{13}$ computation  when predicting quarter resolution flow fields. Moreover, it only contains 1.37M parameters which is as light as SPyNet~\cite{Ranjan_2017}, but behaves much better with $5\times$ speed up, confirming state-of-the-art size-accuracy trade-off. Thanks to the coarse-to-fine structure, FastFlowNet can naturally trade accuracy for speed according to specific robotic applications. For example, our model can process $1024\times436$ resolution images within a range of $5.7$-$26$ FPS on a single  Jetson TX2 GPU. To our %best
knowledge, FastFlowNet represents the first real-time solution for accurate optical flow on embedded devices.
% as shown in Fig~\ref{fig1}.

\section{Related work}

FlowNet~\cite{Fischer_2015} is the first work to use CNNs for optical flow estimation. It propose two variants of FlowNetS and FlowNetC together with the synthetic FlyingChairs dataset for end-to-end training. The %successor
improved version 
FlowNet2~\cite{Ilg_2017} fuses cascaded FlowNets with a small displacement module, and uses FlyingThings3D~\cite{Mayer_2016} dataset for a further fine-tuning schedule. These encoder-decoder based networks have exceeded variational solutions with orders of magnitude faster speed. % however, large model sizes make them unfit for lightweight mobile applications.
However, these models are not sufficiently compact and fast for mobile applications. 

Inspired by the traditional image pyramid, SPyNet~\cite{Ranjan_2017} estimates residual flow in decomposed spatial levels with only 1.2M parameters, % whereas, 
by 
%it suffers from performance degradation. 
sacrificing some performance accuracy. 
Following the coarse-to-fine residual structures, \textit{i.e.}, PWC-Net~\cite{Sun_2018_CVPR} and LiteFlowNet~\cite{Hui_2018_CVPR} employ feature pyramid, warping and correlation in each level to %succeed in raising
achieve good 
accuracy with a reduced number of  parameters, representing the most efficient optical flow architecture. %Following researches are almost exclusively focused on improving accuracy.

Most recent works on optical flow focus on improving accuracy.
IRR-PWC~\cite{Hur_2019} shares the flow decoder and context network among all spatial scales for joint optical flow and occlusion estimation. HD3-Flow~\cite{Yin_2019} decomposes contiguous flow fields into discrete grids and adopts a better backbone. FDFlowNet~\cite{Kong_2020} employs better U-shape network and partial fully connected flow estimators for efficient inference. MaskFlowNet~\cite{zhao2020maskflownet} adds additional occlusion mask module into PWC-Net~\cite{Sun_2018_CVPR} %with
and uses 
cascaded refinement. Recently, RAFT~\cite{teed2020raft} outperforms other approaches by first calculating all-pairs similarity and then performing %a dozen
iterations at a high resolution. %however,
The price is that 
it incurs %huge
significantly larger 
computation burden. 
% 
% YOU HAVE SAID THIS> DO NOT REPEAT
%In short, we should highlight that the efficient LiteFlowNet~\cite{Hui_2018_CVPR} only runs at 1.1 fps when inferring Sintel resolution images on Jetson TX2 GPU, which is not suitable for low-power devices, let alone other heavy models.
We instead  pay more attention to designing a compact model while retaining good optical flow accuracy here. 

\section{Our method}
\subsection{%Approach
Overview of the Approach}

Given two time-adjacent input images $I_1, I_2 \in \mathbb{R}^{H \times W \times 3}$,
our 
proposed FastFlowNet exploits the coarse-to-fine residual structure to estimate gradually refined optical flow $F^{l} \in \mathbb{R}^{H^l \times W^l \times 2}, l=6,5, \ldots, 2$.
But it is extensively reformed to accelerate inference by properly reducing parameters and computation cost. To this end, we first %change
replace 
the dual convolution feature pyramid in PWC-Net~\cite{Sun_2018_CVPR} with the head enhanced pooling pyramid for enhancing high resolution pyramid feature and reducing model size. Then we propose a novel center dense dilated correlation layer for constructing compact cost volume while keeping the large search radius. Finally, new shuffle block decoders are employed at each pyramid level to regress optical flow %with obvious reduction on computation and marginal drops in accuracy.
with significantly cheaper computation.
%
% DO NOT REPEAT
% The overall architecture of FastFlowNet is depicted in Fig~\ref{fig2}, and 
The structure details of the model are listed in Table~\ref{tab1}.

\subsection{Head Enhanced Pooling Pyramid}
Traditional methods~\cite{Bouguet_2000, Brox_2011} apply image pyramid to optical flow estimation for speeding up optimization or dealing with large displacement. SPyNet~\cite{Ranjan_2017} %succeeds to 
transfers this  classical paradigm and first introduces the pooling based image pyramid to deep models. Since raw images are variant to illumination changes and the fixed pooling pyramid is vulnerable to noise~\cite{Brox_2004}, such as shadows and reflection, PWC-Net~\cite{Sun_2018_CVPR} and LiteFlowNet~\cite{Hui_2018_CVPR} replace it by learnable feature pyramids with significant improvement. Concretely, they gradually expands feature channels while shrinking spatial size by half to extract robust matching features.

However, large channel numbers in low resolutions %will occupy lots of 
result in a large number of 
parameters. Moreover, it can be redundant for the coarse-to-fine structure, since low level pyramid features are only responsible to estimate coarse flow fields. Thus, we combine the feature pyramid at  high levels with a pooling pyramid at other lower  levels to obtain both strengths as shown in Fig.~\ref{fig2}. On the other hand, high resolution pyramid features are  relatively shallow in the PWC-Net, as each pyramid level only contains two convolutions with 3$\times$3 kernel that has small receptive field. Therefore,  we add one more convolution %in
at 
high levels to intensify pyramid features  with a small extra cost. By balancing computation among different scales, we have presented a head enhanced pooling pyramid feature extractor as listed in the top of Table~\ref{tab1}. Like FlowNetC, PWC-Net and LiteFlowNet, HEPP generates six pyramid levels from  $\sfrac{1}{2}$ resolution (level 1) to $\sfrac{1}{64}$ resolution (level 6) by a scale factor of 2.

\subsection{Center Dense Dilated Correlation}
One crucial step in modern optical flow architectures is to calculate feature correspondence by the inner product based correlation layer~\cite{Fischer_2015}. Given two pyramid features of ${\rm f}_1^l, {\rm f}_2^l$ in level $l$, like many coarse-to-fine residual approaches, we first adopt the bilinear interpolation based warping operation~\cite{Ranjan_2017, Ilg_2017} to warp the second feature ${\rm f}_2^l$ according to $2\times$ upsampled previous flow field ${\rm up_2}(F^{l+1})$. The warped target feature ${\rm f}_{\rm warp}^l$ can significantly reduce displacement caused by large motions, %that
which 
helps to narrow the search region and simplify the task to estimate relatively small residual flow. Recent models~\cite{Sun_2018_CVPR}\cite{Hui_2018_CVPR} construct cost volume by correlating source features with corresponding warped target features within a local square area that can be formulated as
\begin{equation}
c^l({\bf x}, {\bf d}) = {\rm f}_1^l({\bf x}) \cdot {\rm f}_{\rm warp}^l({\bf x}+{\bf d}) / N, \, {\bf d} \in [-r, r]\times[-r, r]
\label{eq1}
\end{equation}
where $\bf x$ and $\bf d$ represent spatial and offset coordinates. $N$ is the length of input features. $r$ means search radius; and $\cdot$ stands for inner product.

\begin{figure}[t]
	\captionsetup[subfigure]{farskip=1pt}
	\centering
	\subfloat[$r=4$]
	{
		\includegraphics[width=0.3\linewidth]{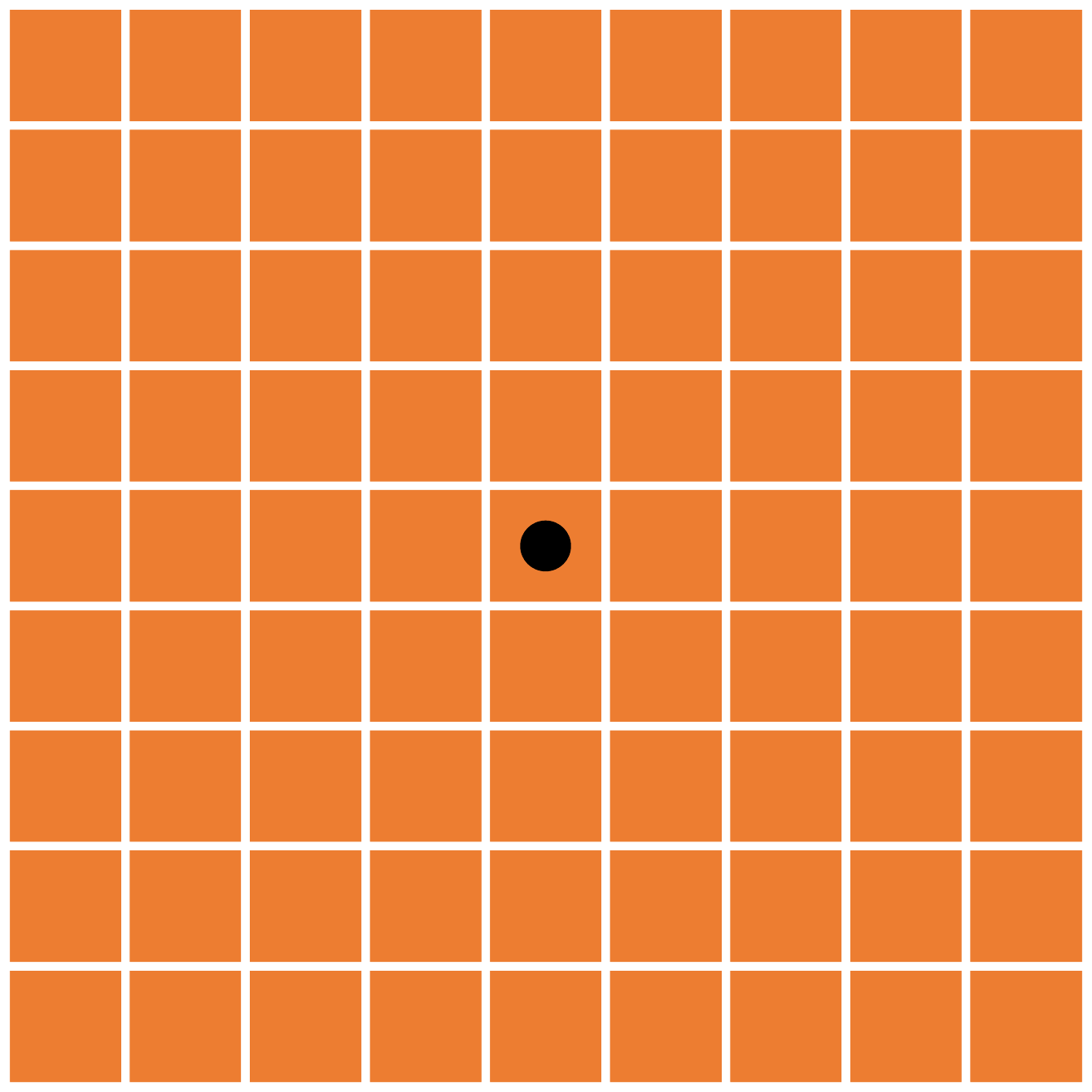}\label{fig3_1}
	}
	\subfloat[$r=3$]
	{
		\includegraphics[width=0.3\linewidth]{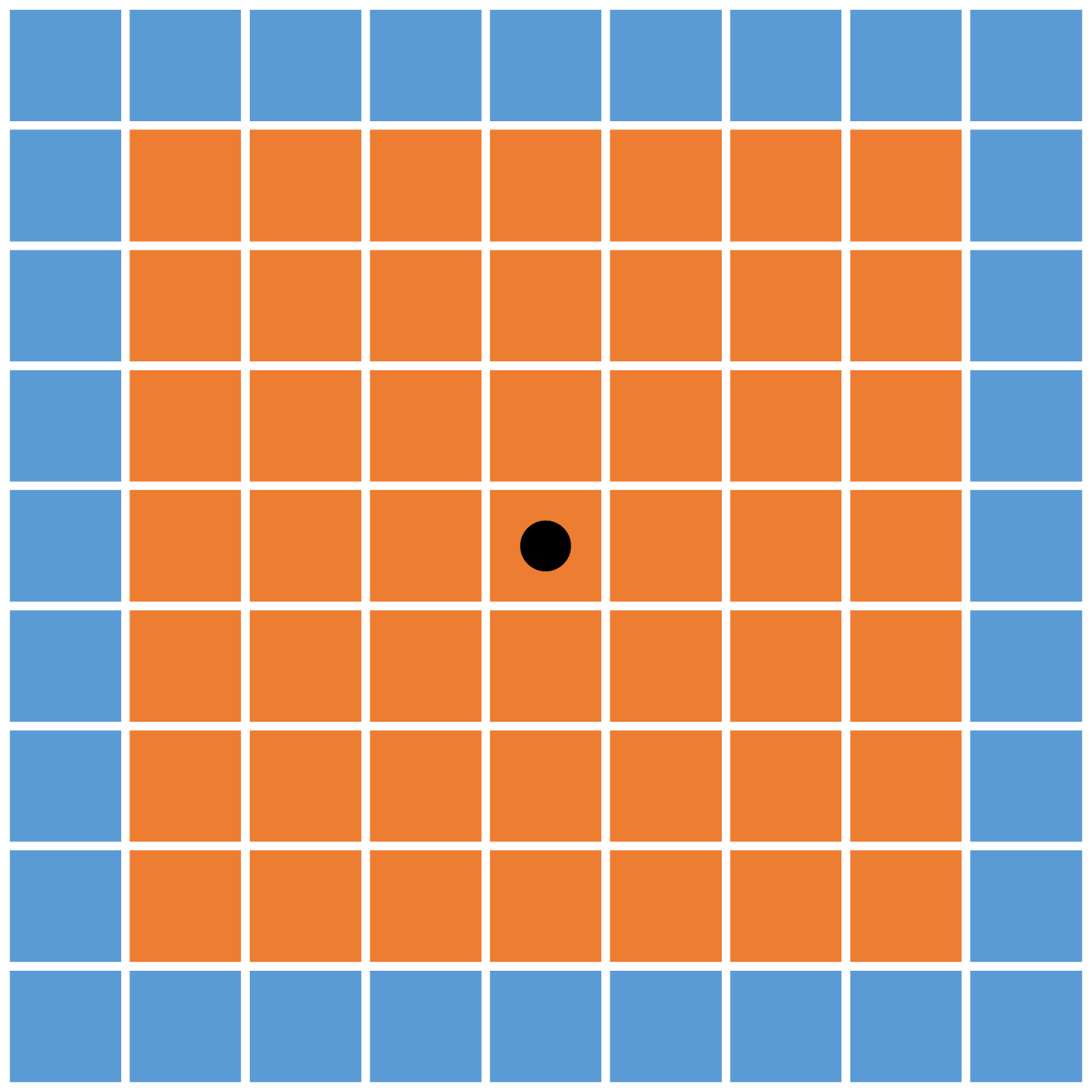}\label{fig3_2}
	}
	\subfloat[CDDC]
	{
		\includegraphics[width=0.3\linewidth]{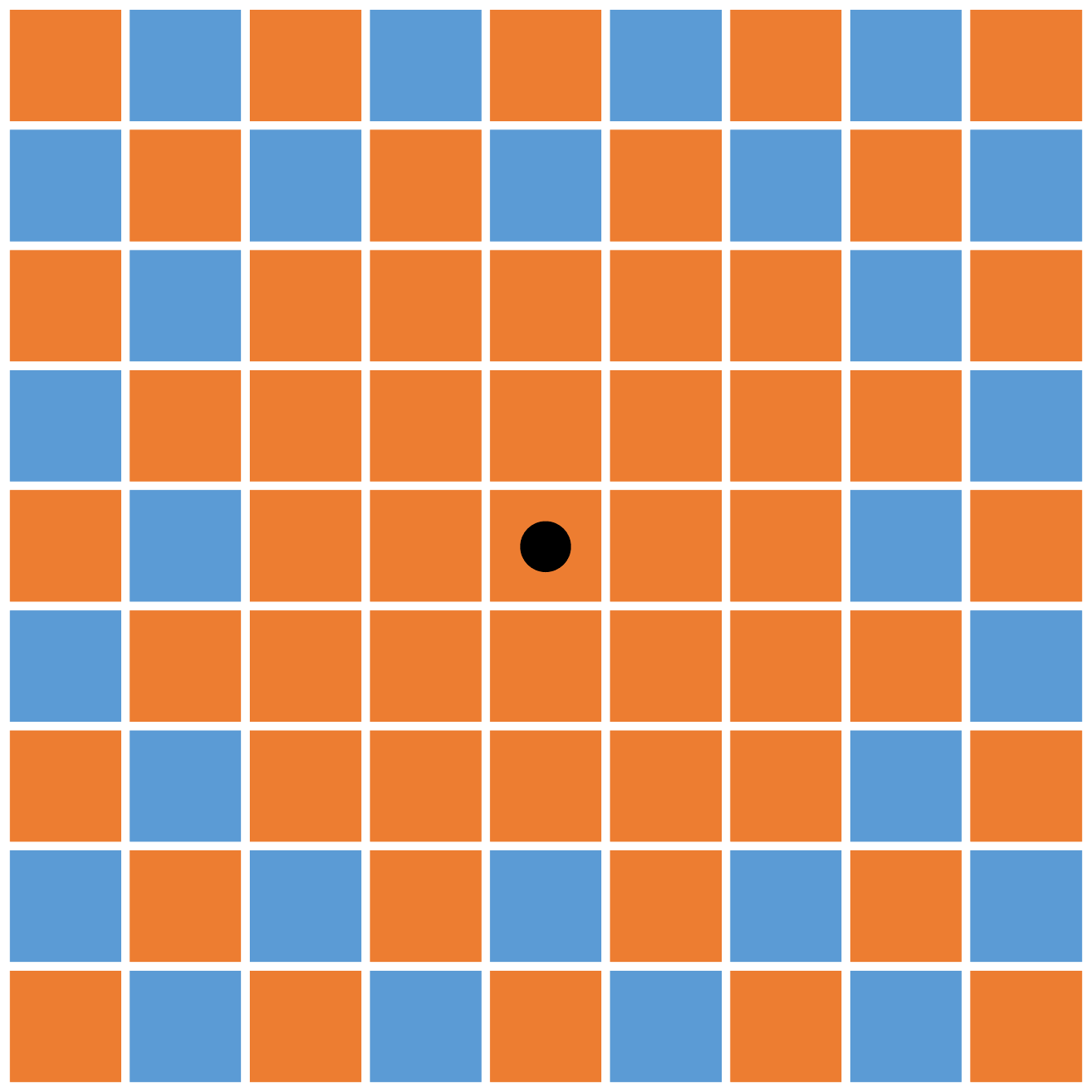}\label{fig3_3}
	}
	\caption{Different approaches for constructing the cost volume. Orange squares represent considered search grids. (a) and (b) are traditional square matching costs while (c) is the new proposed center dense dilated cost volume.}
	\vspace{-5mm}
	\label{fig3}
\end{figure}

Existing works~\cite{Sun_2018_CVPR, hofinger2019improving} have shown that increasing the search radius when building the cost volume can lead to lower end-point error during both training and testing, especially for large displacement cases. However, feature channels of the cost volume are squared %to
in terms of the 
search radius and computation complexity of following decoder network %will be
becomes 
fourth power. As shown in Fig.~\ref{fig3_1}, many flow networks~\cite{Sun_2018_CVPR,Kong_2020,Hur_2019,Yin_2019} set $r = 4$, and the resulting large computation burden has impeded low-power applications. A simple 
way to reduce computation cost is to decrease $r$, such as setting $r = 3$ in Fig~.\ref{fig3_2}. It can reduce cost volume features from 81 to 49, however this method is at the expense of sacrificing perception range and accuracy.

Inspired by the atrous spatial pyramid pooling (ASPP) module in DeepLabv3~\cite{chen2017rethinking}, we propose a novel center dense dilated correlation (CDDC) layer that samples search grids densely around the center while down sampling grid points in large motion areas as shown in Fig.~\ref{fig3_3}. 

Different from ASPP which uses parallel atrous convolutions to capture multi-scale context information, our proposed CDDC is designed to reduce computation when constructing large radius cost volume. In FastFlowNet, it outputs 53 feature channels which has similar computation budget with traditional $r = 3$ setting. Our motivation is that the residual flow distribution is more focused on small motions. Our experiments %will 
show that CDDC behaves better than the naive compression method.

\begin{table}[t]
	\centering
	\large
	\renewcommand\arraystretch{1.06}
	\caption{Architecture of FastFlowNet. HEPP, MFC, SBD represents head enhanced pooling pyramid, multiple feature construction and shuffle block decoder. MFC and SBD share the same structure across all pyramid levels, thus only level 5 are listed for brevity. `Pool' stands for average pooling. Fconv5\_2, fconv5\_3 and fconv5\_4 are convolutions with group $=$3. `Shuffle' means the channel shuffle operation~\cite{8578814}. Convolution layers except upconv6 (deconvolution) and fconv5\_7 are followed by a LeakyReLU activation.}
	\resizebox{0.48\textwidth}{!}{
		\begin{tabular}{ c|c|cc|c|c }
			\hline
			& Layer Name & Kernel & Stride & Input & Ch I/O \\
			\hline
			\multirow{10}{*}{\rotatebox{90}{HEPP}} & pconv1\_1 & 3$\times$3 & 2 & img1 or img2 & 3/16 \\
			~ & pconv1\_2 & 3$\times$3 & 1 & pconv1\_1 & 16/16 \\
			~ & pconv2\_1 & 3$\times$3 & 2 & pconv1\_2 & 16/32 \\
			~ & pconv2\_2 & 3$\times$3 & 1 & pconv2\_1 & 32/32 \\
			~ & pconv2\_3 & 3$\times$3 & 1 & pconv2\_2 & 32/32 \\
			~ & pconv3\_1 & 3$\times$3 & 2 & pconv2\_3 & 32/64 \\
			~ & pconv3\_2 & 3$\times$3 & 1 & pconv3\_1 & 64/64 \\
			~ & pconv3\_3 & 3$\times$3 & 1 & pconv3\_2 & 64/64 \\
			~ & pool4 & 2$\times$2 & 2 & pconv3\_3 & 64/64 \\
			~ & pool5 & 2$\times$2 & 2 & pool4 & 64/64 \\
			~ & pool6 & 2$\times$2 & 2 & pool5 & 64/64 \\
			\hline
			\multirow{5}{*}{\rotatebox{90}{MFC}} & upconv6 & 4$\times$4 & 1/2 & flow6 & 2/2 \\
			~ & warp5 & - & - & pool5b,upconv6 & 64,2/64 \\
			~ & cddc5 & 1$\times$1 & 1 & pool5a,warp5 & 64,64/53 \\
			~ & rconv5 & 3$\times$3 & 1 & pool5a & 64/32 \\
			\cline{3-5}
			~ & concat5 & \multicolumn{3}{|c|}{rconv5 + cddc5 + upconv6} & 32,53,2/87 \\ 
			\hline
			\multirow{10}{*}{\rotatebox{90}{SBD}} & fconv5\_1 & 3$\times$3 & 1 & concat5 & 87/96 \\
			~ & fconv5\_2 & 3$\times$3 & 1 & fconv5\_1 & 3$\times$32/3$\times$32 \\
			~ & shuffle5\_2 & - & - & fconv5\_2 & 3$\times$32/32$\times$3 \\
			~ & fconv5\_3 & 3$\times$3 & 1 & shuffle5\_2 & 3$\times$32/3$\times$32 \\
			~ & shuffle5\_3 & - & - & fconv5\_3 & 3$\times$32/32$\times$3 \\
			~ & fconv5\_4 & 3$\times$3 & 1 & shuffle5\_3 & 3$\times$32/3$\times$32 \\
			~ & shuffle5\_4 & - & - & fconv5\_4 & 3$\times$32/32$\times$3 \\
			~ & fconv5\_5 & 3$\times$3 & 1 & shuffle5\_4 & 96/64 \\
			~ & fconv5\_6 & 3$\times$3 & 1 & fconv5\_5 & 64/32 \\
			~ & fconv5\_7 & 3$\times$3 & 1 & fconv5\_6 & 32/2 \\
			\hline
	\end{tabular}}
	\label{tab1}
\end{table}

\subsection{Shuffle Block Decoder}
After building the cost volume, coarse-to-fine models usually concatenate context features, cost volume and upsampled previous flow as inputs for the following decoder networks. %Existing in 
Due to 
each pyramid level, the decoder part takes up most parameters and computation of the whole network. Thus, better speed-accuracy trade-off is %worth investigation.
critically important.  

PWC-Net~\cite{Sun_2018_CVPR} shows that the dense connected flow decoder can improve accuracy after fine-tuning on the  FlyingThings3D~\cite{Mayer_2016} dataset at the price of increasing the model size and computation. LiteFlowNet~\cite{Hui_2018_CVPR} uses sequential connected flow estimator and also %gets relative good
shows improved 
performance. To achieve better trade-off between these two %topologies,
schemes, 
FDFlowNet~\cite{Kong_2020} employs the partial fully connected structure. However, all these methods can not well support real-time inference on embedded systems.

Thanks to the compact cost volume built by CDDC, we can directly decrease maximal feature channels in decoder network from %previous
128 to current 96 without %convergence 
issues. To further reduce computation and model size, we reform the middle three 96-channel convolutions into group convolutions followed by channel shuffle operations, which we %call it
term `shuffle block'. Different from ShuffleNet~\cite{8578814} as a backbone network, our shuffle block decoder is employed for regressing optical flow. As listed in the bottom of Table~\ref{tab1}, each decoder network contains three shuffle blocks with group $=3$, %that
which 
can efficiently reduce computation with marginal drops in accuracy.

\begin{figure*}[t]
	\centering
	\includegraphics[width=\textwidth]{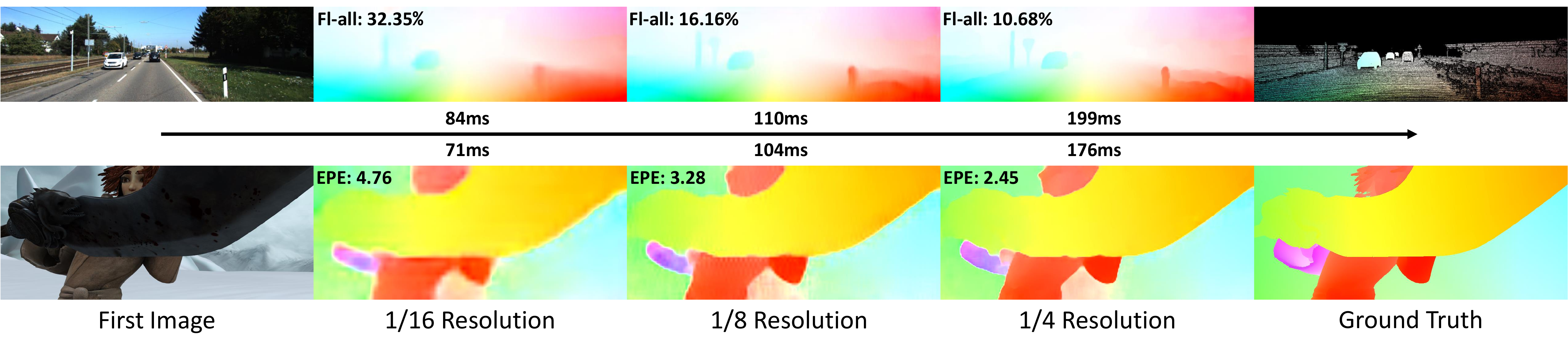}
	\caption{Visual examples of anytime inference. FastFlowNet can naturally trade speed for accuracy according to specific robotic applications. Black arrow means inference timeline which is measured on an  embedded Jetson TX2 GPU.}
	\label{fig4}
\end{figure*}

\begin{table}[t]
	\centering
	\large
	\renewcommand\arraystretch{1.06}
	\caption{Ablation study on structure variants with different pyramid feature extractor, cost volume constructor and optical flow decoder. Flow accuracy is measured by end point error. FLOPs are calculated on Sintel resolution images.}
	\resizebox{0.48\textwidth}{!}{
		\begin{tabular}{ ccc|cc|cc }
			\hline
			Pyramid & Cost & Decoder & Sintel & KITTI & Params.  & FLOPs \\
			\hline
			conv & $r = 4$ & $g = 1$ & 3.12 & 2.37 & 4.23M & 28.2G \\
			conv+pool & $r = 4$ & $g = 1$ & 3.00 & 2.25 & 3.22M & 27.7G \\
			HEPP & $r = 4$ & $g = 1$ & 2.92 & 2.13 & 3.27M & 28.3G \\
			HEPP & $r = 3$ & $g = 1$ & 2.93 & 2.33 & 2.18M & 19.1G \\
			HEPP & CDDC & $g = 1$ & 2.95 & 2.18 & 2.20M & 19.2G \\
			HEPP & CDDC & $g = 2$ & 3.04 & 2.22 & 1.57M & 14.0G \\
			HEPP & CDDC & $g = 3$ & 3.11 & 2.29 & 1.37M & 12.2G \\
			HEPP & CDDC & $g = 4$ & 3.22 & 2.47 & 1.26M & 11.3G \\
			HEPP & CDDC & $g = 6$ & 3.35 & 2.59 & 1.16M & 10.4G \\
			\hline
	\end{tabular}}
	\label{tab2}
\end{table}

\subsection{Loss Function}
Since FastFlowNet has the same pyramid scales as FlowNet~\cite{Fischer_2015} and PWC-Net~\cite{Sun_2018_CVPR}, we adopt the same multi-scale L2 loss for training
\begin{equation}
L_{epe} = \sum_{l=2}^{6} \alpha_{l} \sum_{\bf x} |F^{l}({\bf x}) - F_{gt}^{l}({\bf x})|_{2}
\label{eq2}
\end{equation}
where $|\cdot|_{2}$ computes L2 norm between predicted and ground-truth flow fields. When fine-tuning on datasets with real scene structure, such as KITTI, we use following robust loss
\begin{equation}
L_{robust} = \sum_{l=2}^{6} \alpha_{l} \sum_{\bf x} (|F^{l}({\bf x}) - F_{gt}^{l}({\bf x})| + \epsilon)^{q}
\label{eq3}
\end{equation}
where $|\cdot|$ denotes L1 norm, $\epsilon = 0.01$ is a small constant and $q<1$ makes loss more robust for large magnitude outliers. For fair comparison with previous methods~\cite{Fischer_2015, Ilg_2017, Sun_2018_CVPR, Hui_2018_CVPR}, weights in Eq.~\eqref{eq2} and Eq.~\eqref{eq3} are set to $\alpha_{6}=0.32, \alpha_{5}=0.08, \alpha_{4}=0.02, \alpha_{3}=0.01$ and $\alpha_{2}=0.005$.

\begin{table*}[h!]
	\footnotesize
	\renewcommand\arraystretch{1.1}
	\caption{Performance comparison on benchmark datasets. Speed and computation are measured on Sintel resolution images. %Values
	Results 
	in parentheses are that of networks being trained on the same dataset. Default accuracy metric is the end-point error.}
	%\begin{center}
		\centering
		\resizebox{0.99\textwidth}{!}{
			\begin{tabular}{ r|cc|cc|cc|ccc|cc|cc }
				\hline
				\multirow{2}{*}{Method} & \multicolumn{2}{|c|}{Sintel Clean} & \multicolumn{2}{|c|}{Sintel Final} & \multicolumn{2}{|c | }{KITTI 2012} & \multicolumn{3}{|c|}{KITTI 2015} & Params & FLOPs & \multicolumn{2}{|c }{Time (s)} \\
				& train & test & train & test & train & test & train & train (Fl-all) & test (Fl-all) & (M) & (G) & 1080Ti & TX2 \\
				\hline
				FlowNetC \cite{Fischer_2015} & 4.31 & 7.28 & 5.87 & 8.81 & 9.35 & - & - & - & - & 39.18 & 6055.3 & 0.029 & 0.427 \\
				FlowNetC-ft & (3.78) & 6.85 & (5.28) & 8.51 & 8.79 & - & - & - & - & 39.18 & 6055.3 & 0.029 & 0.427 \\
				FlowNet2 \cite{Ilg_2017} & 2.02 & 3.96 & 3.14 & 6.02 & 4.09 & - & 10.06 & 30.37\% & - & 162.52 & 24836.4 & 0.116 & 1.547 \\
				FlowNet2-ft & (1.45) & 4.16 & (2.01) & 5.74 & (1.28) & 1.8 & (2.30) & (8.61\%) & 11.48\% & 162.52 & 24836.4 & 0.116 & 1.547 \\
				\hline
				SPyNet \cite{Ranjan_2017} & 4.12 & 6.69 & 5.57 & 8.43 & 9.12 & - & - & - & - & 1.20 & 149.8 & 0.050 & 0.918 \\
				SPyNet-ft & (3.17) & 6.64 & (4.32) & 8.36 & (4.13) & 4.7 & - & - & 35.07\% & 1.20 & 149.8 & 0.050 & 0.918 \\
				PWC-Net \cite{Sun_2018_CVPR} & 2.55 & - & 3.93 & - & 4.14 & - & 10.35 & 33.67\% & - & 8.75 & 90.8 & 0.034 & 0.485 \\
				PWC-Net-ft & (2.02) & 4.39 & (2.08) & 5.04 & (1.45) & 1.7 & (2.16) & (9.80\%) & 9.60\% & 8.75 & 90.8 & 0.034 & 0.485 \\
				LiteFlowNet\cite{Hui_2018_CVPR} & 2.48 & - & 4.04 & - & 4.00 & - & 10.39 & 28.50\% & - & 5.37 & 163.5 & 0.055 & 0.907 \\
				LiteFlowNet-ft & (1.35) & 4.54 & (1.78) & 5.38 & (1.05) & 1.6 & (1.62) & (5.58\%) & 9.38\% & 5.37 & 163.5 & 0.055 & 0.907 \\
				\hline
				PWC-Net-small & 2.83 & - & 4.08 & - & - & - & - & - & - & 4.08 & - & 0.024 & - \\
				LiteFlowNetX & 3.58 & - & 4.79 & - & 6.38 & - & 15.81 & 34.90\% & - & 0.90 & - & 0.035 & - \\
				FastFlowNet & 2.89 & - & 4.14 & - & 4.84 & - & 12.24 & 33.10\% & - & 1.37 & 12.2 & 0.011 & 0.176 \\
				FastFlowNet-ft & (2.08) & 4.89 & (2.71) & 6.08 & (1.31) & 1.8 & (2.13) & (8.21\%) & 11.22\% & 1.37 & 12.2 & 0.011 & 0.176 \\
				\hline
		\end{tabular}}
%	\end{center}
	\label{tab3}
\end{table*}

\section{Experiments}
\subsection{Implementation Details}
To compare FastFlowNet with other networks, we follow the two-stage training strategy proposed in FlowNet2~\cite{Ilg_2017}. Ground truth optical flow is divided by 20 and down sampled as supervision at different levels. Since final flow prediction is in a quarter resolution, we use bilinear interpolation to obtain the full resolution optical flow. During both training and fune-tuning, we use the same data augmentation methods as in \cite{Ilg_2017}, including mirror, translate, rotate, zoom, squeeze and color jitter. All our experiments are implemented in PyTorch and conducted on a machine equipped with 4 NVIDIA GTX 1080 Ti GPU cards. To compare performance of different optical flow models on mobile devices, we further evaluate inference %delay
time 
on the embedded Jetson TX2 GPU.

We first train FastFlowNet on FlyingChairs~\cite{Fischer_2015} using $S_{short}$ learning schedule~\cite{Ilg_2017}, \textit{i.e.}, the learning rate is initially set to $1e-4$ and decays half at 300k, 400k and 500k iterations. We crop $320\times448$ patches during data augmentation and adopt a batch size of 8. Then the model is fine-tuned on FlyingThings3D~\cite{Mayer_2016} using the $S_{fine}$ schedule~\cite{Ilg_2017}, \textit{i.e.}, the learning rate is initially set to $1e-5$ and decays by half at 200k, 300k and 400k iterations. Random crop size is $384\times768$ and the batch size is set to 4. The Adam~\cite{Adam_2014} optimizer and multi-scale L2 loss in Eq.~\ref{eq2} are always used. We call the model {\bf FastFlowNet} after above two-stage learning schedules whose results on Sintel and KITTI training sets are listed in Table~\ref{tab3}.

\subsection{Ablation Study}
To explore the efficient optical flow structure with low computation cost, we perform  ablations on the pyramid feature extractor, cost volume constructor and optical flow decoder. Results of the gradually lightened models are %as
listed in Table~\ref{tab2}. All variants are first trained on FlyingChairs and evaluated on Sintel Clean training sets. To test in large displacement cases, which are common when robots move forward or backward, %then 
they are fine-tuned on KITTI 2015 training subset for 5-fold cross-validation. Since change of inference %delay
time 
is not distinct based on our PyTorch implementation, we use FLOPs~\cite{howard2017mobilenets, 8578814} which means number of floating point multiply-add operations to measure real computation complexity. In Table~\ref{tab2}, `conv' represents dual convolution feature pyramid used in PWC-Net, `conv+pool' stands for combining dual convolution feature pyramid in top 3 levels with pooling pyramid in bottom 3 levels. Note that traditional $r = 4$ cost volume has sequential connected flow decoders, whose maximal feature channel is 128 rather than 96, in order to make it fully convergent. $g = 1$ means no grouping convolution without shuffle block in flow decoder.

Surprisingly, the lighter `conv+pool' pyramid feature extractor behaves better than the traditional `conv' method. Our analysis is that parameter-free average pooling may share gradients among different levels during back-propagation, which may have  made pyramid features more robust to scale changes. 

By replacing `conv+pool' with HEPP, we can achieve  better results with small additional overhead. Then, we decrease the search radius for constructing lightened $r = 3$ cost volume to reduce %energy consumption.
computation. 
Flow accuracy on Sintel %has 
shows 
almost no change, while results on large displacement KITTI decline clearly. To mitigate performance degradation, we adopt CDDC to build a compact cost volume with larger receptive field at each pyramid level. Although it does not improve on Sintel training which contains 
%lots of
many 
small movements, it %behaves
works 
better on the more challenging KITTI 2015 validation set. 

Finally, we employ SBD for further compression on parameters and computation. In Table~\ref{tab3}, we gradually lighten the network by setting incremental group numbers in SBD. It can be seen that computation reduction becomes relatively small when $g > 3$, whereas 
%obvious
clear 
performance degradation appears. Thus, we adopt $g = 3$ in FastFlowNet which %gets 
shows a 
better computation-accuracy trade-off.

\subsection{MPI Sintel}

When fine-tuning on the Sintel~\cite{Butler_2012} training sets, we crop $320\times768$ patches and set the batch size to 4. Robust loss in Eq.~\ref{eq3} is adopted with $q = 0.4$. Like PWC-Net, we set the initial learning rate to $5e-5$ and %disturb
change 
it to $3e-5$, $2e-5$ and $1e-5$ every 150k iterations in total 600k iterations,  
%to jump out from local minimum, 
where the learning rate decays to zero in each period. We report results on the Sintel test sets by submitting predictions to the MPI Sintel online benchmark, which are listed in Table~\ref{tab3}. For mobile applications, we also measure computation burden of different networks in FLOPs and real inference %delay
time 
on both desktop GTX 1080Ti GPU and embedded Jetson TX2. All experiments are conducted in PyTorch for fair comparison.

In Table~\ref{tab3}, FastFlowNet outperforms FlowNetC~\cite{Fischer_2015} in all items due to the advanced coarse-to-fine residual structure. It runs $2.5\times$ faster than FlowNetC on both GTX 1080Ti and Jetson TX2 with $500\times$ less computation. The reason why competition complexity is not proportional to inference %delay
time 
is that correlation and warping layers have large memory access cost (MAC) while they are low FLOPs operations.

FastFlowNet outperforms SPyNet~\cite{Ranjan_2017} by $26.4\%$ and $27.3\%$ on Sintel Clean and Final test sets respectively. Although containing %a little 
slightly 
more parameters, our network runs $5\times$ faster with $12\times$ less computation. Compared with the state-of-the-art LiteFlowNet~\cite{Hui_2018_CVPR}, we can approach $90\%$ of its performance with $13.4\times$ less (only 12.2 GFLOPs) computation cost.

To further verify the effectiveness of the proposed flow architecture, % simplicification methods, 
we compare FastFlowNet with %small
compact 
versions of PWC-Net~\cite{Sun_2018_CVPR} and LiteFlowNet~\cite{Hui_2018_CVPR} provided in their papers. PWC-Net-small is the small version of PWC-Net by dropping dense connections in flow decoders;  and LiteFlowNetX is the small variant by shrinking convolution channels in feature pyramid and flow estimators. We can see that FastFlowNet %gets
achieves 
almost the same good results with PWC-Net-small on Sintel training datasets after two-stage pretraining, while being additional $3\times$ samller and $2\times$ faster.

\subsection{KITTI}
To test the proposed architecture on more challenging large displacement real world data, we fine-tune FastFlowNet on mixed KITTI 2012~\cite{Geiger2012CVPR} and KITTI 2015~\cite{7298925} training sets. Like~\cite{Ilg_2017, Sun_2018_CVPR, Hui_2018_CVPR}, we crop $320\times896$ patches and adopt a batch size of 4. The same fine-tuning schedule on Sintel is employed and $q$ in robust loss is set to $0.2$ for outlier suppression. To adapt to driving scenes, spatial augmentation methods of rotate, zoom and squeeze are skipped with a probability of $0.5$ as~\cite{Sun_2018_CVPR, Hui_2018_CVPR}. Then we evaluate the  fine-tuned FastFlowNet on the online KITTI benchmarks which are listed in Table~\ref{tab3}.

Compared with the encoder-decoder based FlowNet2~\cite{Ilg_2017}, our model can achieve %same
\textit{on par }
or better results on KITTI test datasets while being $120\times$ smaller and $10\times$ faster.

We can achieve  similar accuracy %in regard
compared 
to PWC-Net~\cite{Sun_2018_CVPR} with $7.5\times$ less multiply-add computation %burden,
cost. 
%that is of great concern on low-power embedded systems. 
For LiteFlowNetX which simply reduces feature channels, our
proposed HEPP, CDDC and SBD modules work better on all metrics confirming effectiveness of our contribution. Some visual results in anytime setting~\cite{8794003, dovesi2019realtime} are  shown 
in Fig.~\ref{fig4}.

\section{Conclusion}
In this paper, we have proposed a fast and lightweight network for accurate optical flow estimation %towards
for 
mobile applications. Our method is based on the well-known coarse-to-fine structure and is extensively reformed to accelerate inference by properly reducing parameters and computation. The proposed head enhanced pooling pyramid, center dense dilated correlation and shuffle block decoder are efficient modules for pyramid feature extraction, cost volume construction and optical flow decoding, %that 
which 
covers all components of flow estimation pipeline. We have evaluated each component on both synthetic Sintel and real-world KITTI datasets to verify their effectiveness. The proposed FastFlowNet can be directly plugged into any optical flow based robotic vision systems with much less resource consumption,
achieving good accuracy. 
%and nice performance.

%\section*{Acknowledgements}
%This research is % partly
%in part 
%supported by NSFC, China (No: 61876107, U1803261), National Key R\&D Program of China (No. 2019YFB1311503).

%%%%%%%%%%%%%%%%%%%%%%%%%%%%%%%%%%%%%%%%%%%%%%%%%%%%%%%%%%%%%%%%%%%%%%%%%%%%%%%%
\bibliographystyle{ieeetr}
\bibliography{refs}

\end{document}